# Reducing Hallucinations: Enhancing VQA for Flood Disaster Damage Assessment with Visual Contexts

Yimin Sun[1], Chao Wang[2*] and Yan Peng[3]

[1] Shanghai University, Shanghai, China, yimin_sun@shu.edu.cn
[2] Shanghai University, Shanghai, China, *cwang@shu.edu.cn
[3] Shanghai University, Shanghai, China, pengyan@shu.edu.cn

**Abstract.** The zero-shot performance of visual question answering (VQA) models relies heavily on prompts. For example, a zero-shot VQA for disaster scenarios could leverage well-designed Chain of Thought (CoT) prompts to stimulate the model's potential. However, using CoT prompts has some problems, such as causing an incorrect answer in the end due to the hallucination in the thought process. In this paper, we propose a zero-shot VQA named Flood Disaster VQA with Two-Stage Prompt (VQA-TSP). The model generates the thought process in the first stage and then uses the thought process to generate the final answer in the second stage. In particular, visual context is added in the second stage to relieve the hallucination problem that exists in the thought process. Experimental results show that our method exceeds the performance of state-of-the-art zero-shot VQA models for flood disaster scenarios in total. Our study provides a research basis for improving the performance of CoT-based zero-shot VQA.

## 1 Introduction

Damage from natural disasters such as floods is enormous [1]. Getting information on disaster sites helps provide recovery services when a disaster occurs. With the development of Artificial Intelligence, VQA [11] has become an advanced technology for obtaining information in disaster scenarios with its ability to understand images and natural language. Recently, a zero-shot VQA based on flood disaster scenarios has been proposed. Meanwhile, the "out-of-the-box" capability [2] of zero-shot VQA means that it can be directly transferred to other disaster scenarios, thus reducing the time and resource costs of training the model. For example, under a fire disaster, the zero-shot VQA model can be used to learn about smoke conditions or the safety of personnel. And while in an earthquake disaster, zero-shot VQA can be directly used to quickly assess the extent of damage to buildings and the safety of personnel. This rapid transfer capability allows zero-shot VQA to be used in a wide range of disaster scenarios.

Previous zero-shot VQA systems rely heavily on the reasoning capabilities of large language models. All kinds of well-designed prompts can be used to leverage the reasoning potential of these language models [4]. For example, zero-shot VQA for Flood

Disaster Damage Assessment (ZFDDA) is proposed in our previous work [6], which uses the CoT [8] prompt method to stimulate the large model potential. However, there are some problems while using the CoT prompts such as incorrect output. One possible reason is that the model generates answers based on CoT is often divided into two steps (Fig. 1), producing the thought process and outputting the final answer, and the factual bias generated in the thought process often leads to the error of the final answer, which we call the hallucination due to the model's limited capability. This hallucination severely limits the capacity of the model. Thus, the challenge we are facing in this study can be summarised as follows. **Challenge: how to relieve the hallucinations generated by the thought process to improve the accuracy of the zero-shot VQA model?**

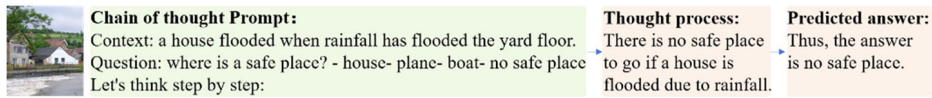

**Fig. 1.** Figure 1.    An inference process of CoT prompt.

To tackle the challenge, inspired by Multimodal-CoT [3], a zero-shot VQA model named Flood Disaster **VQA** with **T**wo-**S**tage **P**rompt (VQA-TSP) is proposed. Given an image and a question, in the first stage, we construct a CoT prompt using the visual context and the question. Then the CoT prompt is used to generate the thought process. In the second stage, visual context is combined with the thought process to create a new prompt, which inspires the model to generate the final answer. Specifically, the introduction of visual context in the second stage is to solve the problem of hallucination during inference caused by CoT prompts.

In summary, to reduce the hallucination produced by CoT prompts in the thought process, visual context is introduced to enhance the model's ability to generate correct answers. **Overall, the main contributions of this research are as follows: (1)** We propose a zero-shot VQA model named VQA-TSP for flood disaster scenario information acquisition. **(2)** This model introduces visual context in the second-stage prompt to address the hallucination problem in the thought process. **(3)** The introduction of visual context can understand and analyze the flood disaster images better, which improves the model's accuracy in answering questions, especially yes-no questions.

## 2    Methods

### 2.1    Problem Definition

The goal of the method is to generate the answer given an image and a question via a two-stage prompt. In the first stage prompt, the CoT prompt is built by the visual context and the question, the output is the thought process, which can be represented as: $r = G_1(i, q)$, where $r$ means the thought process, $i$ means the image, $q$ means the question, and $G_1$ means the thought process generation method. In the second stage, the general prompt consists of visual context and the thought process. The prompt is used to generate the final answer, which can be represented as: $a = G_2(r, v, q)$, Where $a$ means the answer, $r$ means the thought process generated in the last stage, $v$ means the visual context, $q$ means the question, and $G_2$ means our question answering method.

### 2.2 Framework

VQA-TSP model structure (Fig. 2) includes two stages: **(1)** Thought Process Generation and **(2)** Final Answer Generation. Each stage is introduced in detail below.

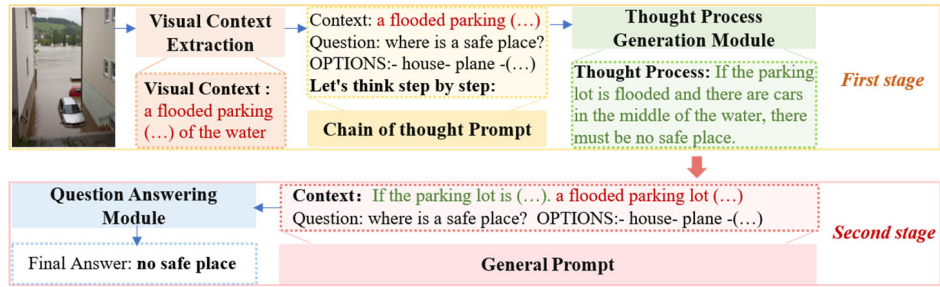

**Fig. 2.** An overview of VQA-TSP model.

**Thought Process Generation.** The task of this stage is to generate a thought process. This stage includes three modules: the Visual Context Extraction module, the Chain of Thought Prompt, and the Thought Process Generation Module. The Visual Context Extraction module is used to extract the visual context that is most related to answering the question [5]. Then the CoT prompt is constructed based on the visual context, the question, and "let's think step by step". After that, the well-designed CoT prompt is delivered into the Thought Process Generation Module. Finally, the model outputs a reliable thought process based on CoT.

**Final Answer Generation.** The task of this stage is to generate the final answer. This stage includes the General prompt and the Question Answering Module. A general prompt is used to inspire the model to output the answer and the prompt is a combination of the thought process generated in the last stage and the visual context. The visual context used is consistent with the visual context used in the last stage. Then the general prompt is delivered into the Question Answering Module to guide the language model to reason out the final answer. Question Answering Module and the Thought Process Generation Module have the same structure. Flan-Alpaca [7] is the backbone of these modules with fewer parameters but with performance comparable to that of GPT-3.5.

## 3 Experiments

### 3.1 Dataset

FFD-IQA [6]: A VQA dataset for flood disaster scenarios contains 2,058 images and 22,422 question-and-answer pairs. This dataset has a wide range of image types and question types. The images are all taken during the flood, and both individuals and buildings in the pictures may suffer varying degrees of damage. The questions in the dataset are mainly focusing on the safety of individuals trapped in disaster sites and the availability of emergency services.

### 3.2 Baseline

**ZFDDA** [6]: A zero-shot VQA model for flood disaster damage assessment. Without pre-training, the VQA model can answer the questions about the flood disaster image. The CoT prompt is the key to stimulating the inference ability and question-answering ability of the model, while the model shows high accuracy on the FFD-IQA dataset compared to other baseline models. However, this model has not yet reached its optimal performance and the limitations of using CoT need to be addressed.

### 3.3 Metric and Evaluation

**Metric.** The Exact Match (EM) metric is a frequently used metric to measure the percentage of predicted answers that exactly match the ground truth answers [10]. However, the output predicted answers of our method are flexible which means those answers do not have fixed content and length and are hard to evaluate by the EM metric. So we define a new metric and make the rules of human evaluation. Accuracy is used as a metric to evaluate the performance of our model. Accuracy is defined as the ratio of the number of plausible answers to the total number of questions. Formally, Accuracy A can be calculated as: $A = \frac{N(p)}{N(q)}$, where $N(p)$ represents the number of plausible answers and $N(q)$ represents the number of total questions. The evaluator's selection criteria and the definition of plausible answers and implausible answers are introduced as follows.

**Human Evaluation.** Volunteer evaluators are chosen following the Inter Annotator Agreement (IAA) [9]. Fleiss' Kappa is used as a metric to evaluate the consistency between evaluators. The final evaluators are selected with a Fleiss' Kappa value of 0.72.

**Scoring Criteria.** The evaluator rates the quality of the output answer based on the scoring criteria. Here, the description of plausible and implausible answers is provided below: **(1) Implausible answer:** An example of an implausible answer is when the answer does not match the content of the image. For example, if your question is "Where is a safe place? -house -plane-boat-no safe place", and then there is a house in the image but the house is flooded. If the answer given by the model is still the house, then the answer is implausible. **(2) Plausible answers:** A plausible example is one where the answer does match the image. For example, if your question is "Where is a safe place? -house -plane-boat-no safe place" and then the image shows the house is flooded, if the model outputs that there is no safe place, then the answer is plausible.

## 4 Results

### 4.1 Accuracy Analysis

Table 1. Accuracy on different VQA settings.

| Methods | All | Multiple-choice | Free-form | Yes-no |
|---|---|---|---|---|
| ZFDDA w/o CoT | 52.06% | 32.05% | 62.18% | 55.03% |
| ZFDDA zero-shot CoT | 57.43% | 33.21% | **83.26%** | 57.36% |
| VQA-TSP (**ours**) | **60.86%** | **34.23%** | 82.15% [a] | **62.70%** |

a. <u>digits</u> represents the second-best result

We run our VQA-TSP on the dataset and the results are as follows. Based on Table 1, the following conclusions can be drawn: **(1)** Compared with the ZFDDA zero-shot CoT model, the overall accuracy of VQA-TSP improves by 3.43%, from 57.43% to 60.86%. In particular, in the yes-no question, VQA-TSP accuracy has the biggest improvement of 5.34% to 62.70%. This shows that the introduction of visual context in the second stage plays a crucial role in improving the accuracy of the VQA task on flood disasters. **(2)** Compared with the ZFDDA zero-shot CoT model, on other types of questions that require logical thinking, a general improvement effect of our model is also observed. Especially in the Multiple-choice questions, VQA-TSP accuracy improves from 33.21% to 34.23%. This shows that the introduction of visual context in the second stage has a positive effect in improving the accuracy of the VQA task on flood disasters. This result also suggests that the thought process generated in the first stage also has an important impact on the model's ability to reason out the correct answer. Therefore, further research needs to take into account the method to improve the accuracy of the thought process to achieve more comprehensive and accurate answer generation.

### 4.2 Qualitative Analysis

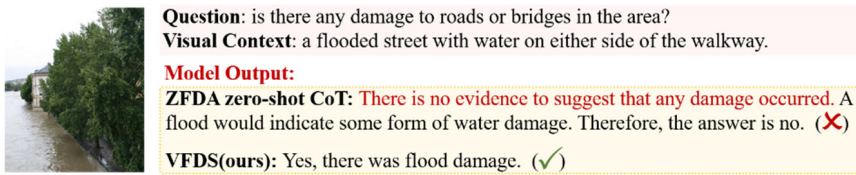

**Fig. 3.** An example of qualitative analysis.

Introducing visual context can relieve the hallucination of the thought process, a qualitative example is shown in Fig. 3. In the image, the question asks "Is there any damage to roads or bridges in the area?", the ZFDDA zero-shot CoT reasons that there is no damage according to the thought process "There is no evidence to suggest that any damage occurred." The answer does not match with the content of the image, which is the reasoning process bias that leads to the wrong final answer. The VQA-TSP model corrects this bias by introducing visual context in the thought process, and our model finally gives the correct answer "There was flood damage".

## 5      Conclusions

In this paper, we propose a Flood Disaster VQA with Two-Stage Prompt (VQA-TSP) model. The framework consists of a two-stage prompt, aiming to utilize a CoT prompt and a general prompt. In particular, combining visual features with the thought process to further stimulate the potential of large models in the second stage is effective when addressing the problem of hallucination in the thought process when using the CoT prompt only. Such a framework combines textual and visual context to improve the model's inference ability, which improves the accuracy of question answering. In the future, research on designing middle steps in the thought process to further prompt our VQA model capability is valuable.

## Acknowledgment

This work was supported by the Program of Natural Science Foundation of Shanghai (No. 23ZR1422800).